\documentclass[10pt, conference]{IEEEtran}

\usepackage{cite}
\ifCLASSINFOpdf
  \usepackage[pdftex]{graphicx}
\else
\fi

\usepackage[cmex10]{amsmath}
\usepackage{array}
\usepackage{url}
\usepackage[utf8]{inputenc}
\usepackage{multicol}
\usepackage{multirow}
\usepackage{amssymb}
\usepackage{color}
\usepackage{balance}
\usepackage{booktabs}
\usepackage{dsfont}
\usepackage{subfig}
\usepackage{algorithm}
\usepackage[noend]{algpseudocode}

\usepackage{algorithm}
\usepackage[noend]{algpseudocode}

\newcommand{\ve}[1]{\mathbf{#1}}
\newcommand{\R}{\mathbb{R}}
\newcommand{\bigO}[1]{\mathcal{O}\left(#1\right)}

\newcounter{LINumberOfComments}
\stepcounter{LINumberOfComments}

\begin{document}
\title{Fast k-NN Search}

\author{\IEEEauthorblockN{
Ville Hyvönen\IEEEauthorrefmark{1}, 
Teemu Pitkänen\IEEEauthorrefmark{1},
Sotiris Tasoulis\IEEEauthorrefmark{2},
Elias Jääsaari\IEEEauthorrefmark{1},
Risto Tuomainen\IEEEauthorrefmark{1},\\
Liang Wang\IEEEauthorrefmark{3},
Jukka Corander\IEEEauthorrefmark{1}\IEEEauthorrefmark{4}\IEEEauthorrefmark{5} 
and Teemu Roos\IEEEauthorrefmark{1}
}
\IEEEauthorblockA{\IEEEauthorrefmark{1}Helsinki Institute for Information Technology HIIT, 
University of Helsinki, Finland. Email: \{firstname.lastname\}@cs.helsinki.fi}
\IEEEauthorblockA{\IEEEauthorrefmark{2}Liverpool John Moores University, UK. Email: S.Tasoulis@ljmu.ac.uk}
\IEEEauthorblockA{\IEEEauthorrefmark{3}Computer Laboratory, University of Cambridge,
UK. Email: liang.wang@cl.cam.ac.uk}
\IEEEauthorblockA{\IEEEauthorrefmark{4}Pathogen Genomics, Wellcome Trust Sanger Institute, Cambridge, UK}
\IEEEauthorblockA{\IEEEauthorrefmark{5}Department of Biostatistics, University of Oslo,  Norway. Email: jukka.corander@helsinki.fi}
}

%\author{\IEEEauthorblockN{Sotiris Tasoulis}
%\IEEEauthorblockA{Helsinki Institute for Information Technology HIIT\\
%Department of Computer Science \\
%University of Helsinki, Finland\\
%tasoulis@cs.helsinki.fi}
%
%\and \IEEEauthorblockN{Teemu Pitkänen}
%\IEEEauthorblockA{HIIT\\
%Department of Computer Science \\
%University of Helsinki, Finland\\
%teemu.pitkanen@helsinki.fi}
%
%\and \IEEEauthorblockN{Ville Hyvönen}
%\IEEEauthorblockA{HIIT\\
%Department of Computer Science \\
%University of Helsinki, Finland\\
%ville.o.hyvonen@helsinki.fi}
%}


\maketitle

\begin{abstract}
Efficient index structures for fast approximate nearest neighbor queries are required in many applications such as recommendation systems. In high-dimensional spaces, many conventional methods suffer from excessive usage of memory and slow response times. We propose a method where multiple random projection trees are combined by a novel voting scheme. The key idea is to exploit the redundancy in a large number of candidate sets obtained by independently generated random projections in order to reduce the number of expensive exact distance evaluations. The method is straightforward to implement using sparse projections which leads to a reduced memory footprint and fast index construction. Furthermore, it enables grouping of the required computations into big matrix multiplications, which leads to additional savings due to cache effects and low-level parallelization. We demonstrate by extensive experiments on a wide variety of data sets that the method is faster than existing partitioning tree or hashing based approaches, making it the fastest available technique on high accuracy levels.

%Random projection trees have proven to be effective for approximate nearest neighbor searches in high-dimensional spaces where conventional methods are not applicable due to excessive usage of memory and computational time. We show that growing multiple trees on the same data can improve the performance even further, without increasing the total computational cost of queries when the tree parameters are controlled in a suitable way. Our experiments identify optimal parameter values to achieve accurate searches with extremely fast query times. In empirical experiments, the proposed algorithm achieves a better tradeoff between speed and accuracy than existing state-of-the-art methods.
%\liang{I think we need to emphasise the voting in the abstract since it is an important contribution imo. Something like "We also propose a voting scheme to generalise the standard MPRT algorithm, which not only reduces computation complexity but also improves accuracy ..."}

\end{abstract}

\begin{IEEEkeywords}
Nearest Neighbor Search; Random Projections; High Dimensionality; Approximation Algorithms
\end{IEEEkeywords}

% \fi
\IEEEpeerreviewmaketitle

\section{Introduction}
\label{sec:intro}

Nearest neighbor search is an essential part of many machine learning algorithms. Often, it is also the most time-consuming stage of the procedure, see~\cite{Muja2014}. In applications areas, such as in recommendation systems, robotics and computer vision, where fast response times are critical, using brute force linear search is often not feasible. This problem is further magnified by the availability of inreasingly complex, high-dimensional data sources. 

Applications that require frequent $k$-NN queries from large data sets are for example object recognition \cite{Nister2006, Lowe2004}, shape recognition \cite{Amit1997} and image completion \cite{Hays2007} using large databases of image descriptors,  and content-based web recommendation systems \cite{Wang2016}.

Consequently, there is a vast body of literature on the algorithms for fast nearest neighbor search. These algorithms can be divided into exact and approximate nearest neighbor search. While  exact nearest neighbor search algorithms return the true nearest neighbors of the query point, they suffer from the curse of dimensionality: their performance degrades when the dimension of the data increases, rendering them no better than brute force linear search in the high-dimensional regime. 

In approximate nearest neighbor search, the main interest is the tradeoff between query time and accuracy, which can be measured either in terms of distance ratios or the probability of finding the true nearest neighbors. Several different approaches have been proposed and their implementations are commonly used in practical applications.

In this section, we first define the approximate nearest neighbor search problem, then review the existing approaches to it, and finally outline a new method that avoids the main drawbacks of existing methods.

\subsection{Approximate nearest neighbor search}

%The problem we consider here is approximate nearest neighbor search. One formulation is $(1 + \epsilon)$--approximate $k$-NN search \cite{Arya1998}, in which the distance of the points returned by the algorithm is at most $(1 + \epsilon)$ times the distance of the true $k$th nearest neighbor of the query point. However,  in this article we consider a slightly different version of the problem: we do not set any error bounds, but assess the efficiency of the algorithms empirically by examining the $k$-NN query time required to reach a given accuracy level.

Consider a metric space $\mathcal{M}$, a data set $\ve{X} = (\ve{x}_1, \dots, \ve{x}_n) \subseteq \mathcal{M}$, a query point $\ve{q} \in \mathcal{M}$, and a distance metric $m: \mathcal{M}^2 \rightarrow \R$. The $k$-nearest neighbor ($k$-NN) search is the task of finding the $k$ closest (w.r.t. $m$) points to $\ve{q}$ from the data set $\ve{X}$, i.e., find a set $\ve{K} \subseteq \ve{X}$ for which it holds that $|\ve{K}| = k$ and
\[
m(\ve{q},\ve{x}) \leq m(\ve{q},\ve{y})
\]
for all $\ve{x} \in \ve{K}$, $\ve{y} \in \ve{X} \setminus \ve{K}$.

In this work, we consider applications where $\mathcal{M}$ is the $d$-dimensional Euclidean space $\R^d$, and $m$ is Euclidean distance 
\[
m(\ve{x},\ve{y}) = \sqrt{\sum_{i=1}^d (x_i-y_i)^2}.
\]

An approximate nearest neighbor search algorithm can be divided into two separate phases: an offline phase, and an online phase. In the offline phase an index is built, and the online phase refers to completing fast nearest neighbor queries using the constructed index. In practical applications, the most critical considerations are the accuracy of the approximation (the definition of the accuracy and the required accuracy level depending on the application), and the query time needed to reach it.

As a measure of accuracy, we use recall, which is relevant especially for information retrieval and recommendation settings. It is defined as the proportion of true $k$-nearest neighbors of the query point returned by the algorithm:
\[
\mathrm{Recall} = \frac{|\ve{A} \cap \ve{K}|}{k}.
\]
Here $\ve{A}$ is a set of $k$ approximate nearest neighbors returned by the algorithm, and $\ve{K}$ is the set of true $k$ nearest neighbors of the query point.

Of secondary importance are the index construction time in the offline stage and the memory requirement, which are usually not visible to the user but must be feasible. Note that in many applications the index must be updated regularly when we want to add new data points into the set from where the nearest neighbors are searched from. 

\subsection{Related work}

Most effective methods for approximate nearest neighbor search in high-dimensional spaces can be classified into either hashing, graph, or space-partitioning tree based strategies.

\subsubsection{Hashing based algorithms}

The most well-known and effective hashing based algorithms are variants of locality-sensitive hashing (LSH) \cite{Indyk1998, Gionis1999, Andoni2006}. In LSH, several hash functions of the same type are used. These hash functions are \emph{locality-sensitive}, which means that nearby points fall into the same hash bucket with higher probability than the points that are far away from each other.  When answering a $k$-NN query, the query point is hashed with all the hash functions, and then a linear search is performed in the set of data points in the buckets where the query point falls into. Multi-probe LSH \cite{Lv2007, Dong2008, Andoni2015} improves the efficiency of LSH by searching also the nearby hash buckets of the bucket a query point is hashed into. Thus a smaller amount of hash tables is required to obtain a certain accuracy level. The choice of the hash function affects the performance and needs to be done carefully in each case. 

\subsubsection{Graph based algorithms}

Approximate graph based methods such as \cite{Hajebi2011, Dong2011, Wang2012} build a $k$-NN-graph of the data set in an offline phase, and then utilize it to do fast $k$-NN queries. For example in \cite{Wang2012} several graphs of random subsets of the data are used to approximate the exact $k$-NN-graph of the whole data set. The data set is divided hierarchically and randomly into subsets, and the $k$-NN-graphs of these subsets are built. This process is then repeated several times, and the resulting graphs are merged and the final graph is utilized to answer nearest neighbor queries. Graph based methods are in general efficient, but the index construction is slow for larger data sets.

\subsubsection{Space-partitioning trees}

The first space partitioning-tree based strategy proposed for nearest neighbor search was the $k$-d tree \cite{Bentley1975}, which divides the data set hierarchically into cells that are aligned with the coordinate axes. Nearest neighbors are then searched by performing a backtracking or priority search in this tree. $k$-d trees and other space-partitioning trees can be utilized in approximate nearest neighbor search by terminating the search when a predefined number of data points is checked, or all the points within some predefined error bound from the query point are checked \cite{Arya1998}. 

Instead of the hyperplanes, clustering algorithms can be utilized to partition the space hierarchically; for example the $k$-means algorithm is used in $k$-means trees \cite{Fukunaga1975, Muja2014}. Other examples of clustering based partitioning trees are cover trees \cite{Beygelzimer2006}, VP trees \cite{Yianilos1993} and ball trees \cite{Leibe2006}. Similarly to graph-based algorithms, clustering based variants have the drawback of long index construction times due to hierarchical clustering on large data sets.

An efficient scheme to increase the accuracy of space-partitioning trees is to utilize several parallel randomized space-partitioning trees. Randomized $k$-d trees \cite{Silpa2008, Muja2009} are grown by choosing a split direction at random from dimensions of the data in which it has the highest variance. Nearest neighbor queries are answered using priority search; a single priority queue, which is ordered by distance of the query point from splitting point in the splitting direction, is maintained for all the trees. 

Another variant of randomized space-partitioning tree is the random projection tree (RP tree), which was first proposed for vector quantization in \cite{Dasgupta2009}, and later modified for approximate nearest neighbor search in \cite{Dasgupta2013}. In random projection trees the splitting hyperplanes are aligned with the random directions sampled from the unit sphere instead of the coordinate axes. Nearest neighbor queries are answered using \emph{defeatist search}: first the query point is routed down in several trees, and then a brute force linear search is performed in the union of the points of all the leaves the query point fell into.

%However, we introduce here both a more efficient way to randomize space-partitioning trees, and a more efficient way utilize them in a nearest neighbor search.

%A promising approach to the nearest neighbor search is the use of random projection trees (RP trees). They were first proposed for vector quantization in \cite{Dasgupta2009}, and later modified for approximate nearest neighbor search in \cite{Dasgupta2013}. Random projection tree is a form of space-partitioning tree resembling the classic $k$-d tree. In random projection trees the splitting hyperplanes are aligned with the random directions sampled from the unit sphere instead of the coordinate axes. 

%The intuition behind using random projections is that projecting the data onto a lower dimensional subspace by multiplying it by on orthogonal random matrix can be used for dimensionality reduction as effectively as PCA, but with much lower computational cost \cite{Bingham2001, Achlioptas2001}. On the other hand the randomness inherent in the tree growing process enables growing many independent trees with different sets of random vectors and combining the results; this can be utilized to reach very high accuracy when a sufficiently high number of trees are used. A strong point of random projection trees is also their relative simplicity: they have only two tuning parameters, number and depth of the trees used, and the method is quite robust with respect to these parameters.

\subsection{MRPT algorithm}

One of the strengths of randomized space-partitioning trees is in their relative simplicity compared to hashing and graph based methods: they have only few tunable parameters, and their performance is quite robust with respect to them.
They are also perfectly suitable for parallel implementation because the trees are independent, and thus can be easily parallelized either locally or over the network with minimal communication overhead. 

However, both of the aforementioned variants of randomized space-partitioning trees have a few weak points which limit their efficiency and scalability.  First, randomization used in randomized $k$-d trees does not make the trees sufficiently decorrelated to reach high levels of accuracy efficiently. Randomization used in RP trees is sufficient, but it comes at high computational cost: computation of random projections is time-consuming for high-dimensional data sets because the random vectors have the same dimension as the data. 

Second, both the priority queue search used in \cite{Silpa2008, Muja2009} and the defeatist search used in \cite{Dasgupta2013} require checking a large proportion of the data points to reach high accuracy levels; this leads to slow query times. Third, because each node of an RP tree has its own random vector, the number of random vectors required is exponential with respect to tree depth. In the high-dimensional case this on the one hand increases the memory required by the index unnecessarily, and on the other hand slows down the index construction process.

Because of the first two reasons the query times of the randomized space-partitioning trees are not competitive with the fastest graph based methods (cf. experimental results in  Section \ref{sec:results}).

In this article we propose a method that uses multiple random projection trees (MRPT); it incorporates two additional features that lead to fast query times and accurate results. The algorithm has the aforementioned strengths of randomized space-partitioning trees but avoids their drawbacks. 

More specifically, our contributions are:

\begin{enumerate}
\item We show that sparse random projections can be used to obtain a fast variant of randomized space-partitioning trees.

\item We propose the MRPT algorithm that combines multiple trees by \emph{voting search} as a new and more efficient method to utilize randomized space-partitioning trees for approximate nearest neighbor search.

\item We present a time and space complexity analysis of our algorithm.

\item We demonstrate experimentally  that the MRPT algorithm with sparse projections and voting search outperforms the state-of-the-art methods using several real-world data sets across a wide range of sample size and dimensionality.

%\item We present the multiple random projection tree (MRPT) algorithm that is optimized for high dimensionality and large sample sizes, and which can be easily and efficiently parallelized.
%\item We show experimentally how increasing the number of trees increases the number of true nearest neighbors found, even while the size of the set where a final linear search is performed stays constant or even decreases.
%\item We demonstrate the efficiency of the algorithm for approximate nearest neighbor search by comparing its performance to the state-of-the-art methods. 
\end{enumerate}

%An efficient space partitioning-tree based scheme of Muja and Lowe \cite{Muja2009, Muja2014} utilizes randomised k-d trees, which are grown by choosing a split direction at random among the five dimensions along which the data set has a highest variance. Nearest neighbor queries are answered using priority search; a single priority queue, that is ordered by distance of the query point from splitting point in the splitting direction, is maintained for all the trees. 
%{\bf what are the drawbacks with this approach?}

%Each of the existing approaches has limits in terms of the tradeoff between search accuracy and query time and other performance characteristics. High-dimensional data, high accuracy, and fast response times are still critical for many practical applications. In this paper, we propose methods that consistently achieve better performance than the aforementioned approaches especially in the high dimensional, high accuracy regime.

%\section{Random projection trees}
%\label{sec:RP_tree}

In the following sections we describe the proposed method by first revisiting the classic RP tree method~\cite{Dasgupta2013}, and then describing the novel features that lead to increased accuracy and reduced memory footprint and query time.

\section{Index construction}
\label{sec:growing}

\subsection{Classic random projection trees}

% An RP tree is a form of space-partitioning tree. It resembles a k-d tree, but the splitting hyperplanes are aligned with random directions instead of the coordinate axes. 

%We construct the index for the MRPT algorithm by growing $T$ different RP trees as follows.

At the root node of the tree, the points are projected onto a random vector $\ve{r}$ generated from the $d$-dimensional standard normal distribution $N_d(\ve{0}, \ve{I})$. The data set is then split into two child nodes using the median of the points in the projected space: points whose projected values are less or equal to the median in the projected space are routed into the left child node, and points with projected values greater than the median are routed into the right child node.  So the points 
\[
\ve{x}^{(1)}, \dots , \ve{x}^{(\lceil n/2 \rceil)}
\]
fall into the left branch, and the points
\[ 
\ve{x}^{(\lceil n/2 \rceil + 1)} , \dots , \ve{x}^{(n)} 
\]
fall into the right branch. Here $\ve{x}^{(1)}, \dots , \ve{x}^{(n)}$ denotes the ordering of the points $\ve{x}_1, \dots , \ve{x}_n$  with respect to the values of their projections $\ve{x}_1^T \ve{r}, \dots , \ve{x}_n^T \ve{r}$. This process is then repeated recursively for the subsets of the data in both of the child nodes, until a predefined depth $\ell$ is met. 

%{\bf TR: strictly speaking, we claim later that the number of points in different leaf nodes differs at most by one --- however, the above strategy where the left child always gets at least as many points as the right child, leads to a bit more imbalanced distribution of the points. if we do it as above in the implementation, we should in principle modify the computational complexity analysis... however, it may be just be better to just forget this as it is really a minor issue.} 

For high-dimensional data, the computation of random projections is slow, and the memory requirement for storing the random vectors is high. To reduce both the computation time and the memory footprint, we propose two improvements: 
\begin{itemize}
\item Instead of using dense vectors sampled from the $d$-dimensional normal distribution, we use sparse vectors to reduce both storage and computation complexity.
\item Instead of using different projection vectors for each intermediate node of a tree, we use one projection vector for all the nodes on the same level of a tree, so that we can further reduce the storage requirement and maximize low-level parallelism through vectorization.
\end{itemize}

In addition, instead of splitting at a fractile point chosen at random from the interval $[\frac{1}{4}, \frac{3}{4}]$, as suggested in~\cite{Dasgupta2013}, we split at the median to make the performance more predictable and to enable saving the trees in a more compact format. 

In the following subsections we briefly discuss the details of the above improvements. 
Pseudocode for constructing the index using sparse RP trees is given in detail in Algorithms \ref{algo:1}--\ref{algo:2} below.
The proposed Algorithm \ref{algo:1} consists of three embedded \textbf{for} loops. The outermost loop (line 4 - 15) builds $T$ RP trees by continuously calling the \texttt{GROW\_TREE} function in Algorithm \ref{algo:2}. For each individual RP tree, the two inner \textbf{for} loops (line 6 - 12) prepare a sparse random matrix $\ve{R}$ which will be used for projecting the data set $\ve{X}$ to $\ve{P}$ using $\ell$ random vectors (at line 13). Specifically, the innermost loop is for constructing a $d$-dimensional sparse vector for a given tree level.
In Algorithm \ref{algo:2}, the \texttt{GROW\_TREE} function constructs a binary search tree by recursively splitting the high-dimensional space $\ve{X}$ into sub-spaces using the previous projection $\ve{P}$.

\begin{algorithm}
\begin{algorithmic}[1]
\Function{grow\_trees}{$\ve{X}$, $T$, $\ell$, $a$}
  \State n $\leftarrow$ $\ve{X}$.nrows
  \State let trees[$1 \dots T$] be a new array
  \For{$t$ in $1, \dots, T$}
    %\State generate $\ve{r}_{t,1}, \dots, \ve{r}_{t,\ell}$ from $N_d(\ve{0}, \ve{I})$
    \State let $\ve{R}$ be a new $d \times \ell$ matrix
	\For{level in $1, \dots , \ell$}
		\For{$i$ in $1, \dots , d$}
			\State generate $z$ from Bernoulli($a$)
			\If{$z=1$}
            \State generate $\ve{R}$[$i$, level] from $N(0,1)$
			\Else
            \State  $\ve{R}$[$i$, level] $\leftarrow 0$ 
			\EndIf
		\EndFor
	\EndFor    
    \State $\ve{P}$ $\leftarrow$ $\ve{X} \ve{R}$
    \State trees[$t$].root $\leftarrow$ \Call{grow\_tree}{$\ve{X}$, [$1 \dots n$], 0, $\ve{P}$}
    \State trees[$t$].random\_matrix $\leftarrow$ $\ve{R}$
  \EndFor
  \State \textbf{return} trees
\EndFunction
\end{algorithmic}
\caption{Grow $T$ RP trees.}
\label{algo:1}
\end{algorithm}

\begin{algorithm}
\begin{algorithmic}[1]
    \Function{grow\_tree}{$\ve{X}$, indices, level, $\ve{P}$}
  \If{level $= \ell$}  
  \State return $\ve{X}$[indices, ] as leaf
  \EndIf
  \State proj $\leftarrow$ $\ve{P}$[indices, level]
  \State split $\leftarrow$ median(proj)
  \State left\_indices $\leftarrow$ indices[proj $\leq$ split]
  \State right\_indices $\leftarrow$ indices[proj $>$ split]
  \State left $\leftarrow$ \Call{grow\_tree}{$\ve{X}$, left\_indices, level + 1, $\ve{P}$}
  \State right $\leftarrow$ \Call{grow\_tree}{$\ve{X}$, right\_indices, level + 1, $\ve{P}$}
  \State \textbf{return} split, left, right
\EndFunction
\end{algorithmic}
\caption{Grow a single RP tree.}
\label{algo:2}
\end{algorithm}

% \begin{algorithm}
% \begin{algorithmic}[1]
% \Function{grow\_trees}{$\ve{X}$, $T$, $\ell$, $a$}
%   \State let trees[$1 \dots T$] be a new array
%   \For{$t$ in $1, \dots, T$}
%     %\State generate $\ve{r}_{t,1}, \dots, \ve{r}_{t,\ell}$ from $N_d(\ve{0}, \ve{I})$
% 	\For{level in $1, \dots , \ell$}
% 		\For{$i$ in $1, \dots , d$}
% 			\State generate $z$ from Bernoulli($a$)
% 			\If{$z=1$}
% 				\State generate $\ve{r}_{t,\mathrm{level},i}$ from $N(0,1)$
% 			\Else
% 				\State  $\ve{r}_{t,\mathrm{level},i} \leftarrow 0$ 
% 			\EndIf
% 		\EndFor
% 	\EndFor    
%     \State trees[$t$] $\leftarrow$ \Call{grow\_tree}{$\ve{X}$, 0, $\ve{r}_t$}
%   \EndFor
%   \State \textbf{return} trees
% \EndFunction
% \end{algorithmic}
% \caption{Grow $T$ RP trees.}
% \label{algo:1}
% \end{algorithm}

% \begin{algorithm}
% \begin{algorithmic}[1]
% \Function{grow\_tree}{$\ve{X}$, level, $\ve{r}_t$}
%   \If{level $= \ell$}  
%     \State return $\ve{X}$ as leaf
%   \EndIf
%   \State proj $\leftarrow \ve{X} \ve{r}_{t,\mathrm{level}}$
%   \State split $\leftarrow$ median(proj)
%   \State left $\leftarrow$ \Call{grow\_tree}{$\ve{X}$[proj $\leq$ split, ], level + 1, $\ve{r}_t$}
%   \State right $\leftarrow$ \Call{grow\_tree}{$\ve{X}$[proj $>$ split, ], level + 1, $\ve{r}_t$}
%   \State \textbf{return} split, left, right
% \EndFunction
% \end{algorithmic}
% \caption{Grow a single RP tree.}
% \label{algo:2}
% \end{algorithm}

\subsection{Sparse random projections}

With high-dimensional data sets, computation of the random vectors easily becomes a bottleneck on the performance of the algorithm. However, it is not necessary to use random vectors sampled from the $d$-dimensional standard normal distribution to approximately preserve the pairwise distances between the data points. Achlioptas  \cite{Achlioptas2001} shows that the approximately distance-preserving low-dimensional embedding of Johnson-Lindenstrauss-lemma is obtained also with sparse random vectors with components sampled from $\{-1, 0, 1\}$ with respective probabilities $\{\frac{1}{6}, \frac{2}{3}, \frac{1}{6}\}$. Li \emph{et al.} \cite{Li2006} prove that the same components with respective probabilities $\{\frac{1}{2\sqrt{d}}, 1-\frac{1}{\sqrt{d}}, \frac{1}{2\sqrt{d}}\}$, where $d$ is the dimension of the data, can be used to obtain a $\sqrt{d}$-fold speed-up without significant loss in accuracy compared to using normally distributed random vectors. 

We use sparse random vectors $\ve{r} = (r_1, \dots , r_d)$, whose components are sampled from the standard normal distribution with probability $a$, and are zeros with probability $1-a$ :
\[
r_i = 
\begin{cases}
N(0,1) \quad &\text{with probability}\,\, a \\
0 \quad &\text{with probability}\,\, 1-a.
\end{cases}
\]

The sparsity parameter $a$ can be tuned to optimize performance but we have observed that $a = \frac{1}{\sqrt{d}}$ recommended in \cite{Li2006} tends to give near-optimal results in all the data sets we tested, which suggests that further fine-tuning of this parameter is unnecessary. This proportion is small enough to provide significant computational savings through the use of sparse matrix libraries. 

\subsection{Compactness and speed with fewer vectors}

In classic RP trees, a different random vector is used at each inner node of a tree, whereas we use the same random vector for all the sibling nodes of a tree. This choice does not affect the accuracy at all, because a query point is routed down each of the trees only once; hence, the query point is projected onto a random vector $\ve{r}_i$ sampled from the same distribution at each level of a tree. This means that the query point is projected onto i.i.d.\ random vectors $\ve{r_1, \dots , r_\ell}$ in both scenarios. 

An RP tree has $2^\ell - 1$ inner nodes; therefore, if each node of a tree had a different random vector as in classic RP trees, $2^\ell - 1$ different random vectors would be required for one tree. However, when a single vector is used on each level, only $\ell$ vectors are required. This reduces the amount of memory required by the random vectors from exponential to linear with respect to the depth of the trees.

Having only $\ell$ random vectors in one tree also speeds up the index construction significantly. While some of the observed speed-up is explained by a decreased amount of the random vectors that have to be generated, mostly it is due to enabling the computation of all the projections of the tree in one matrix multiplication: the projected data set $\ve{P} \in \R^{n \times \ell}$ can be computed from the data set $\ve{X} \in \R^{n \times d}$ and a random matrix $\ve{R} \in \R^{d \times \ell} $ as 
\[
\ve{P} = \ve{X} \ve{R}.
\]
Although the total amount of computation stays the same, in practice this speeds up the index construction significantly due to the cache effects and low-level parallelization through vectorization.

\subsection{Time and space complexity: Index construction}

At each level of each tree the whole data set is projected onto a $d$-dimensional random vector that has on average $ad$ non-zero components, so the expected\footnote{The following complexity results hold exactly, if the algorithm is modified so that each random vector has exactly $\lceil ad \rceil$ non-zero components instead of the expected number of non-zero components being $ad$.} index construction time is $\Theta(T\ell n ad)$. For classic RP trees ($a=1$) this is $\Theta(T\ell n d)$, but for sparse trees ($a = \frac{1}{\sqrt{d}}$) this is only $\Theta(T\ell n \sqrt{d})$. 

For each tree, we need to store the points allocated to each leaf node, so the memory required by the index is at least $\bigO{Tn}$. At each node only a split point is saved; this does not increase the space complexity because there are only $2^\ell - 1 < n$ inner nodes in one tree.

The expected amount of memory required  by one random vector is $\Theta(ad)$ when random vectors are saved in sparse matrix form. This is $\Theta(d)$ for dense RP trees, and $\Theta(\sqrt{d})$ for sparse RP trees with the sparsity parameter fixed to $a=\frac{1}{\sqrt{d}}$. Because an RP tree has $2^\ell - 1$ inner nodes, the memory requirement for $T$ classic RP trees, which have a different random vector for each node of a tree, is $\bigO{Tdn}$. However, in our version, in which a single vector is used on each level, there are only $\ell$ vectors; hence, the memory requirement for $T$ sparse RP trees is $\bigO{T(\sqrt{d}\log n + n)}$.

\section{Query phase}

In many approximate nearest neighbor search algorithms the query phase is further divided into two steps: a candidate generation step, and an exact search step. 

In the candidate generation step, a candidate set $S$, for which usually $|S| \ll n$, is retrieved from the whole data set, and then in the exact search step $k$ approximate nearest neighbors of a query point are retrieved by performing a brute force linear search in the candidate set. In the MRPT algorithm, the candidate generation step consists of traversal of $T$ trees grown in the index construction phase.

The leaf to which a query point $\ve{q}$ belongs to is retrieved by first projecting $\ve{q}$ at the root node of the tree onto the same random vector as the data points, and then assigning it into the left or right branch depending on the value of the projection. If it is smaller than or equal to the cutpoint $s$ (median of the data points belonging to that node in the projected space) saved at that node, i.e. 
\[
\ve{q}^T \ve{r}_i \leq s, 
\] 
the query point is routed into the left child node, and otherwise into the right child node. This process is then repeated recursively until a leaf is met. 

The query point is routed down into a leaf in all the $T$ trees obtained in the index construction phase. The query process is thus far similar to the one described in \cite{Dasgupta2013}. The principal difference is the candidate set generation: in classic RP trees, the candidate set $S$ includes all the points that belong to the same leaf with the query point in at least one of the trees.  
%For a given query point $\ve{q}$ and its $i$:th nearest neighbor $x_{(i)}$, the events
%\[
%\{f_t(x_{(i)}; \ve{q})=0\}, \, t \in \{1, \dots, T\}
%\]
%that they end up into different leaves in $t$:th tree, are independent because the probability is over the random vectors, which are i.i.d. Hence, the probability that the algorithm fails to return $x_{(i)}$ is 
%\[
%P\{f(x_{(i)}; \ve{q})=0\} = \prod_{i=1}^n P\{f_t(x_{(i)}; \ve{q})=0\}. 
%\] 
A problem with this approach is that when a high number of trees are used in the tree traversal step, the size of the candidate set $|S|$ becomes excessively large. 

In the following, we show how the frequency information (i.e., how frequently a point falls into the same cell as query point) can be utilized to improve both query performance and accuracy.

\subsection{Voting search}

Assume that we have constructed $T$ RP trees of depth $\ell$. Each of them partitions $\R^d$ into $2^\ell$ cells (leaves) $L_1, \dots L_{2^\ell}$, all of which contain $\lceil \frac{n}{2^\ell} \rceil$ or $\lfloor \frac{n}{2^\ell} \rfloor$ data points. For $1\leq t\leq T$, let $f_t$ be an indicator function of data point $\ve{x} \in \ve{X}$ and the query $\ve{q}$, which returns $1$, if $\ve{x}$ and $\ve{q}$ reside in the same cell in tree $t$, and $0$ otherwise:
\[
f_t(\ve{x};\ve{q}) = \sum_{m=1}^{2^\ell} \mathds{1}\{\ve{x} \in L_m, \ve{q} \in L_m\}.
\]
Further, let $F$ be a count function of data point $\ve{x}$, which returns the number of trees in which $\ve{x}$ and $\ve{q}$ belong to the same leaf:
\[
F(\ve{x};\ve{q}) = \sum_{t=1}^T f_t(\ve{x};\ve{q}).
\]

We propose a simple but effective \emph{voting search} where we choose into the candidate set only data points residing in the same leaf as the query point in at least $v$ trees:
\[
S = \{\ve{x} \in \ve{X}: F(\ve{x};\ve{q}) \geq v \}.
\]
The vote threshold $v$ is a tuning parameter. A lower threshold value yields higher accuracy at the expense of increased query times.

This further pruning of the candidate set utilizes the intuitive notion that the closer the data point is to the query point, the more probably an RP tree divides the space so that they both belong to the same leaf. %We do not yet have any theoretical proof for this, but our experimental results show the superiority of this approach.
We emphasize that our voting scheme is not restricted to RP trees, but it can be used in combination with any form of space-partitioning algorithms, given that there is enough randomness involved in the process to render the partitions sufficiently independent.  

Pseudocode for the online stage of the MRPT algorithm is given in detail in Algorithms \ref{algo:3}--\ref{algo:4} below (the knn($\ve{q},k,S$) function is a regular $k$-NN search which returns $k$ nearest neighbors for the point $\ve{q}$ from the set $S$).

\begin{algorithm}
\begin{algorithmic}[1]
\Function{tree\_query}{$\ve{q}$, tree}
  \State $\ve{R}$ $\leftarrow$ tree.random\_matrix
  \State $\ve{p}$ $\leftarrow$ $\ve{q}^T \ve{R}$
  \State root $\leftarrow$ tree.root

  \For{level in $1, \dots, \ell$}
    \If{$\ve{p}$[level] $\leq$ root.split}
      \State root $\leftarrow$ root.left
    \Else
      \State root $\leftarrow$ root.right
    \EndIf
  \EndFor
  \State \textbf{return} data points in root
\EndFunction
\end{algorithmic}
\caption{Route a query point into a leaf in an RP tree.}
\label{algo:3}
\end{algorithm}

\begin{algorithm}
\begin{algorithmic}[1]
\Function{approximate\_knn}{$\ve{q}$, $k$, trees, $v$}
  \State $S \leftarrow \emptyset$
  \State let votes[$1 \dots \ve{X}\text{.nrows}$] be a new array
  \For{tree in trees}
    \For{point in \Call{tree\_query}{$\ve{q}$, tree}}
      \State votes[point] $\leftarrow$ votes[point] + 1
      \If {votes[point] $= v$}
        \State $S \leftarrow S \,\, \cup \{$point$\}$
      \EndIf
    \EndFor
    % \State $\ve{X} \leftarrow$ $\ve{X} \,\, \cup$ tree\_query($\ve{q}$, $t$, $\ell$)
  \EndFor
  \State \textbf{return} knn($\ve{q}, k, S$)
\EndFunction
\end{algorithmic}
\caption{Approximate $k$-NN search using multiple RP trees.}
\label{algo:4}
\end{algorithm}

\subsection{Time and space complexity: Query execution}

%\liang{I guess we can still call this section "Time and space complexity". Although voting does not help in reducing space complexity, we can still ad one sentence to clarify this.}
When the trees are grown into some predetermined depth $\ell$ using the median split, the expected running time of the tree traversal step is $\Theta(T\ell ad)$\footnote{Adding the leaf points  into the candidate set $S$ for each $T$ trees adds a term $\bigO{T \frac{n}{2^\ell}}$ into the time complexity of the tree traversal phase for defeatist search. Because checking the vote count is a constant time operation this term is the same for voting search.  However, in both cases these terms are dominated by the exact search in the search set, which is an $\bigO{T \frac{n}{2^\ell} d}$ operation, so they are not taken into account here.}, because at each level of each RP tree the query point is projected onto a $d$-dimensional random vector that has on average $ad$ non-zero components. When using random vectors sampled from the $d$-dimensional standard normal distribution, this is $\Theta(T\ell d)$, but when using value $a = \frac{1}{\sqrt{d}}$ for the sparsity parameter, it is only $\Theta(T\ell\sqrt{d})$. 

When the median split is used, each leaf has either $\lceil \frac{n}{2^\ell} \rceil$ or $\lfloor \frac{n}{2^\ell} \rfloor$ data points. Therefore, the size of the final search set satisfies $|S| \leq T \lceil \frac{n}{2^\ell} \rceil$ for both defeatist and voting search. Increasing the vote threshold $v$ decreases the size of the search set $S$ in practice, which explains the significantly reduced query times when using voting search (cf. experimental results in Section \ref{sec:results}). However, because the effect of $v$ on $|S|$ depends on the distribution of the data, a detailed analysis of the time complexity requires further assumptions on the data source and is not addressed here. Without any assumptions, both defeatist search and voting search have the same theoretical upper bound for $|S|$. 

Because of this upper bound for the size of the candidate set, the running time of the exact search in the candidate set is $\bigO{T \frac{n}{2^\ell} d}$ for both defeatist search and voting search. Thus, the total query time is $\bigO{Td(\ell + \frac{n}{2^\ell})}$ for classic RP trees, and $\bigO{Td(\frac{\ell}{\sqrt{d}} + \frac{n}{2^\ell})}$ for the MRPT algorithm.

\begin{table}[tb!]
%\begin{footnotesize}
\begin{center}
\caption{Time and space complexity of the algorithm compared to classic RP trees.}
\label{table:complexity}
\begin{tabular}{lll}
\toprule
  & RP trees & MRPT $\left(a = \frac{1}{\sqrt{d}}\right)$\\
\midrule
%Candidate generation time & $\Theta(T\ell d)$ & $\Theta(T\ell\sqrt{d})$ \\ 
%Final search time& $\bigO{T \frac{n}{2^\ell} d}$ & $\bigO{T \frac{n}{2^\ell} d}$ \\
Query time & $\bigO{Td(\ell + \frac{n}{2^\ell})}$ & $\bigO{Td(\frac{\ell}{\sqrt{d}} + \frac{n}{2^\ell})}$ \\
Index construction time & $\Theta(T\ell nd)$ & $\Theta(T\ell n\sqrt{d})$ \\
Index memory & $\bigO{Tdn}$ & $\bigO{T(\sqrt{d}\log n + n)}$ \\
\bottomrule 
\end{tabular}
\end{center}
%\end{footnotesize}
\end{table}

\section{Experimental results}
\label{sec:results}

We assess the efficiency of our algorithm by comparing its against the state-of-the-art approximate nearest neighbor search algorithms. To make the comparison as practically relevant as possible, we chose widely used methods that are available in optimized libraries. % and are known to perform well in benchmarks. {\bf TR: cite for example the Spotify benchmarks and any others that show that these are the best competitors.} 

All of the compared libraries, including ours, are implemented in C++ and compiled with similar optimizations. We make our comparison and the implementation of our algorithm available as open-source\footnote{\url{https://github.com/ejaasaari/mrpt-comparison}}.

All of the experiments were performed on a single computer with two Intel Xeon E5540 2.53GHz CPUs and 32GB of RAM. No parallelization beyond that achieved by the use of linear algebra libraries, such as caching and vectorization of matrix operations, was used in any of the experiments.

\subsection{Data sets}
We made the comparisons using several real-world data sets (see Table \ref{table:datasets}) across a wide range of sample size and dimensionality. The data sets used are typical in applications in which fast nearest neighbor search is required. 
For all data sets, we use a disjoint random sample of 100 test points to evaluate the query time and recall.

The BIGANN and GIST data sets\footnote{\url{http://corpus-texmex.irisa.fr/}} consist of local 128-dimensional SIFT descriptors and global 960-dimensional color GIST descriptors extracted from images \cite{Jegou2011A, Jegou2011B}, respectively. As the SIFT data set, we use a random sample of $n = 2500000$ data points from the BIGANN data set.

In addition to feature descriptor data sets, we consider three image data sets for which no feature extraction has been performed: 1) the MNIST data set\footnote{\url{http://yann.lecun.com/exdb/mnist/}} consists of $28 \times 28$ pixel grayscale images each of which is represented as a vector of dimension $d=784$; 2) the Trevi data set\footnote{\url{http://phototour.cs.washington.edu/patches/default.htm}} consists of $64 \times 64$ pixel grayscale image patches represented as a vector of dimension $d=4096$ \cite{Winder2007}; and the STL-10 data set\footnote{\url{http://cs.stanford.edu/~acoates/stl10}}, which is an image recognition data set containing $96 \times 96$ pixel images from 10 different classes of objects \cite{Coates2011}.

%The audio data set is a sample of features extracted from English sentences spoken by different speakers. It was used as a test data set in \cite{Dong2008}, and further information about the preprocessing steps of the data set can be found there. The data set is available at \url{http://www.cs.princeton.edu/cass/audio.tar.gz}. 

%For word vectors, distance between vectors can be used to measure semantic similarity of the corresponding words. The GloVe data sets\footnote{\url{http://nlp.stanford.edu/projects/glove/}} consist of word vectors trained from different corpora \cite{Pennington2014}. We used the data set pre-trained from tweets consisting of 200-dimensional vectors.

We also use a news data set that contains web pages from different news feeds converted into a term frequency-inverse document frequency (TF-IDF) representation. Dimensionality of the data is reduced to $d=1000$ by applying latent semantic analysis (LSA) to the TF-IDF data. More elaborate description of the preprocessing of the news data set is found in \cite{Wang2016}. 

Finally, we include results for a synthetic data set comprised of $n = 50000$ random vectors sampled from the 4096-dimensional standard normal distribution and normalized to unit length.

\begin{table}[tb!]
%\begin{footnotesize}
\begin{center}
\caption{Data sets used in the experiments}
\label{table:datasets}
\begin{tabular}{l l l l}
\toprule
Data set & $n$ & $d$ & type \\
\midrule
%GloVe & 1193514 & 200 \\
GIST & 1000000 & 960 & image descriptors \\
SIFT & 2500000 & 128 & image descriptors\\
MNIST & 60000 & 784 & image \\
Trevi & 101120 & 4096 & image\\
STL-10 & 100000 & 9216 & image\\
News & 262144 & 1000 & text (TF-IDF + LSA)\\
Random & 50000 & 4096 & synthetic\\
\bottomrule 
\end{tabular}
\end{center}
%\end{footnotesize}
\end{table}

\subsection{Compared algorithms}

% We consider only recall levels over 0.8 since these are the most relevant for the applications of approximate nearest neighbor search.

We compare the MRPT algorithm to  representatives from all three major groups of approximate nearest neighbor search algorithms: tree-, graph- and hashing-based methods.

To test the efficiency of our proposed improvements to classic RP trees, we include our own implementation\footnote{The version of RP trees used in the comparisons is already a slightly improved version: we used RP trees with the same random vector at every level and median split (cf. section \ref{sec:growing}). If we had used a different random vectors at each node of the tree we would have ran out of memory on the larger data sets. However, we emphasize that using the same random vector at every level actually slightly decreases the query times while keeping the accuracy intact; hence, the real performance gap between RP trees and the MRPT algorithm is actually even wider than shown in the results.} of classic RP trees suggested in \cite{Dasgupta2013}, and an MRPT-algorithm with vote threshold $v=1$, which amounts to using sparse RP trees without the voting scheme.

Of the tree-based methods we include randomized $k$-d trees \cite{Silpa2008} and hierarchical $k$-means trees \cite{Muja2009}. Authors of \cite{Muja2014} suggest that these two algorithms give the best performance on data sets with a wide range of dimensionality. Both of the algorithms are implemented in FLANN\footnote{\url{http://www.cs.ubc.ca/research/flann/}}.

%For k-d trees, we use the grid method to search for the optimal combinations of points to check ($c$) and number of trees to use ($T$). For hierarchical k-means trees, we similarly search for the optimal $c$ and branching factor $b$. We set the number of k-means iterations as 15.

We also use as a benchmark the classic $k$-d tree algorithm~\cite{Bentley1975} modified for approximate nearest neighbor search as described in~\cite{Arya1998}; it is implemented in the ANN library\footnote{\url{http://www.cs.umd.edu/~mount/ANN/}}.

% We use standard search with different values for the error bound $\epsilon$, and the sliding midpoint split rule as suggested by the authors.

Among the graph-based methods, the NN-descent algorithm suggested in~\cite{Dong2011} is generally thought to be a leading method. It first constructs a $k$-NN graph using local search in the offline phase, and then utilizes this graph for fast $k$-NN queries in the online phase. The algorithm is implemented in the KGraph library\footnote{\url{http://www.kgraph.org}}.

%We search for the optimal parameters by varying the cost parameter $P$ and the number of neighbors for each node in the constructed graph ($K$).

As a representative of hashing-based solutions we include a multi-probe version of cross-polytope LSH suggested in~\cite{Andoni2015} as implemented in the FALCONN library\footnote{\url{http://www.falconn-lib.org}}. All these algorithms and implementations were selected because they represent the state-of-the-art in their respective categories.

% To tune the algorithm, we search for the optimal number of hash tables and hash functions to use.

For the most important tuning parameters (as stated by the authors), we use grid search on the appropriate ranges to find the optimal parameter combinations in the experiment, or the default values whenever provided by the authors. 
The performance of the MRPT algorithm is controlled by the number of trees, the depth of the trees, the sparsity parameter $a$, and the vote threshold. We fix the sparsity parameter as $a = 1/\sqrt{d}$, and optimize the other parameters using the same grid search procedure as for the other methods. For ANN, we vary the error bound $\epsilon$, and for FLANN, we control the accuracy primarily by tuning the maximum number of leafs to search, but also vary the number of trees for $k$-d trees and the branching factor for hierarchical $k$-means trees. KGraph is primarily controlled by the search cost parameter, but we also build the graph with varying number of nearest neighbors for each node. For FALCONN, we search for the optimal number of hash tables and hash functions. The supplementary material on GitHub provides the exact parameter ranges of grid search as well as interactive graphs that show the optimal values for each obtained result.

% For the most important tuning parameters (as stated by authors), we use grid search on the appropriate ranges to find the optimal parameter combinations. For the other parameters, we use values recommended by the authors.  The exact parameter values used can be found on our comparison code available on github. {\bf TR: do we provide any sensitivity analyses on github in the form of results for other than optimal parameter settings? If yes, this should be mentioned here.}

% Multi-probe LSH implemented in the LSHKIT C++ library,\footnote{\url{http://lshkit.sourceforge.net/}} is an efficient version of LSH~\cite{Dong2008}. We use automatic tuning system provided with the library to tune the number of hash functions $M$ and width of the projection $W$. We set the number of hash tables to $L=10$, and the number of points to check per hash table to $T=100$ which seemed to provide consistently efficient performance with different data sets and values of $k$.

\subsection{Results}

In the experiments we used three different values of $k$ that cover the range used in typical applications: $k=1, 10$ and $100$. However, due to space restrictions, we present results for multiple values of $k$ only for MNIST, which is representative of the general trend. For the other six data sets we only provide results for $k=10$. (The results for $k=1,100$ are available in the supplementary material on GitHub).

Figure~\ref{fig:comparison} shows results for $k = 10$ on six data sets. The times required to reach a given recall level are shown in Table \ref{table:results}. The MRPT algorithm is significantly faster than other tree-based methods and LSH on all of the data sets, expect SIFT. It is worth noting that SIFT has a much lower dimension ($d=128$) than the other data sets. These results suggest that the proposed algorithm is more suitable for high-dimensional data sets.

\begin{figure*}[!ht]
\captionsetup[subfigure]{labelformat=empty}
\centering
\includegraphics[width=\textwidth]{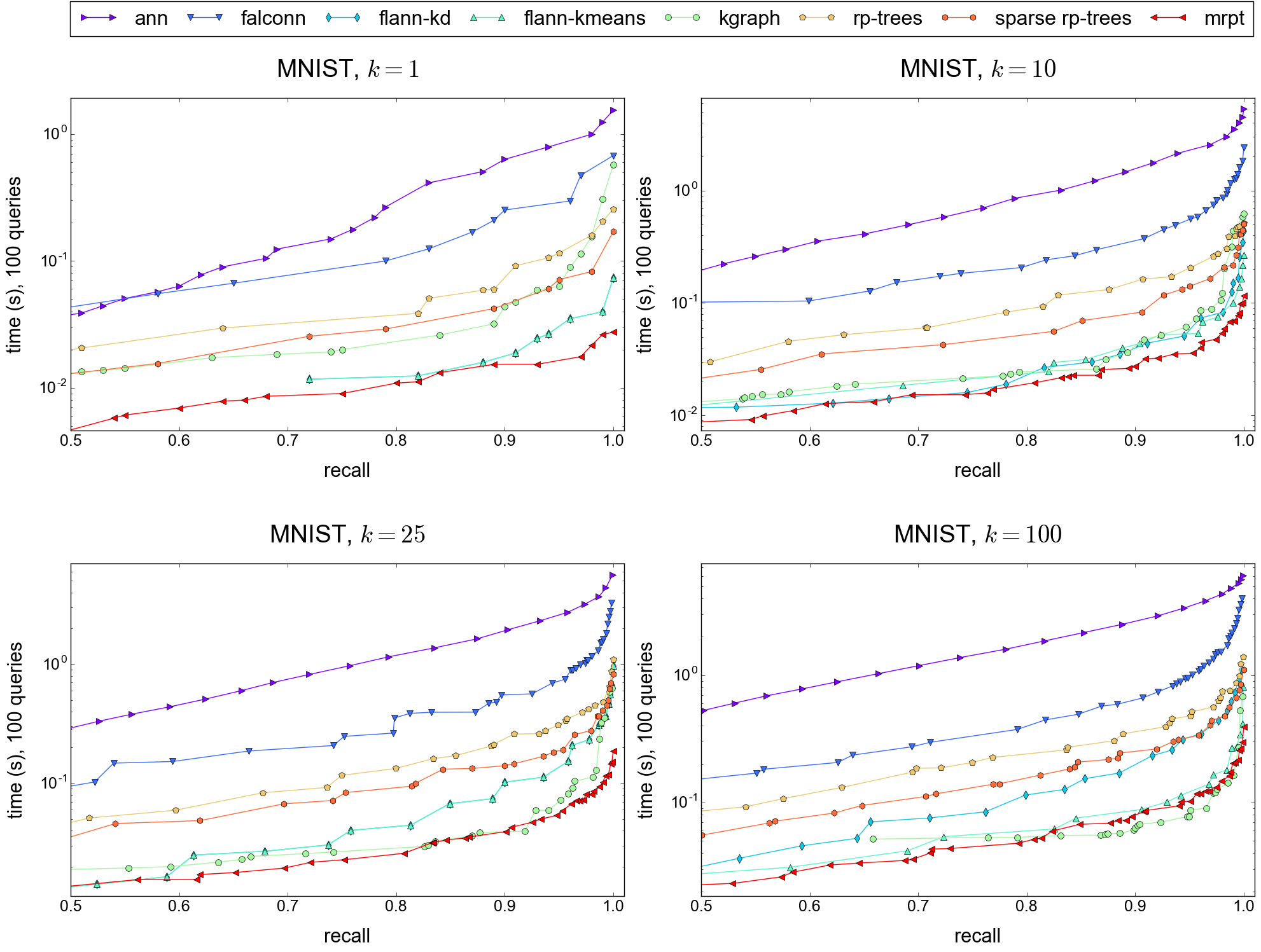}
\caption{Results for $k \in \{1,10,25,100\}$ on the MNIST data set.}
\label{fig:diff_k}
\end{figure*}

For the MNIST data set, results for all three values of $k$ (and additionally for $k=25$) are shown in Figure~\ref{fig:diff_k}.   For the other data sets the results for $k=1 $ and $k=100$ follow a similar pattern as the results for the MNIST data set: the choice of $k$ has no significant effect to the relative performance of the other methods, but the relative performance of KGraph seems to degrade significantly for small values of $k$. 

KGraph is the fastest method on low recall levels, but its performance degrades rapidly on high recall levels (over 90\%), thus making the MRPT algorithm the fastest method on very high recall levels (over 95\%) on all the data sets except SIFT. Our hypothesis is that the flexibility obtained on the one hand by using completely random split directions, and on the other hand by using a high number of trees, which is enabled by sparsity and voting, allows the MRPT algorithm to obtain high recall levels for high-dimensional data sets very efficiently.  

The performance of the MRPT algorithm is also superior to classic RP trees on all of the data sets. In addition, from our two main contributions, voting seems the be more important than sparsity with respect to query times: sparse RP trees are only slightly faster than dense RP trees, the gap being somewhat larger on the higher-dimensional data sets, but the voting search provides a marked improvement on all data sets. This shows that the voting search, especially when combined with sparsity, is an efficient way to reduce query time without compromising accuracy.

The numerical values in Table \ref{table:results} indicate that for recall levels $\geq$ 90\%, the MRPT method is the fastest in 14 out of 21 instances, while the KGraph method is fastest in 7 out 21 cases. Compared to brute force search, MRPT is roughly 25--100 times faster on all six real-world data sets  at 90\% recall level, and roughly 10--40 times faster even at the highest 99\% recall level.

\section{Conclusion}

We propose the multiple random projection tree (MRPT) algorithm for approximate $k$-nearest neighbor search in high dimensions. The method is based on combining multiple sparse random projection trees using a novel voting scheme where the final search is focused to points occurring most frequently among points retrieved by multiple trees. The algorithm is straightforward to implement and exploits standard fast linear algebra libraries by combining calculations into large matrix--matrix operations. 

We demonstrate through extensive experiments on both real and simulated data that the proposed method is faster than state-of-the-art space-partitioning tree and hashing based algorithms on a variety of accuracy levels. It is also faster than a leading graph based algorithm (KGraph) on high accuracy levels, while being slightly slower on low accuracy levels. The good performance of MRPT is especially pronounced for high-dimensional data sets.

Due to its very competitive and consistent performance, and simple and efficient index construction stage --- especially compared to graph-based algorithms --- the proposed MRPT method is an ideal method for a wide variety of applications where high-dimensional large data sets are involved.

\begin{figure*}[hbtp]
\captionsetup[subfigure]{labelformat=empty}
\centering
\includegraphics[width=\textwidth]{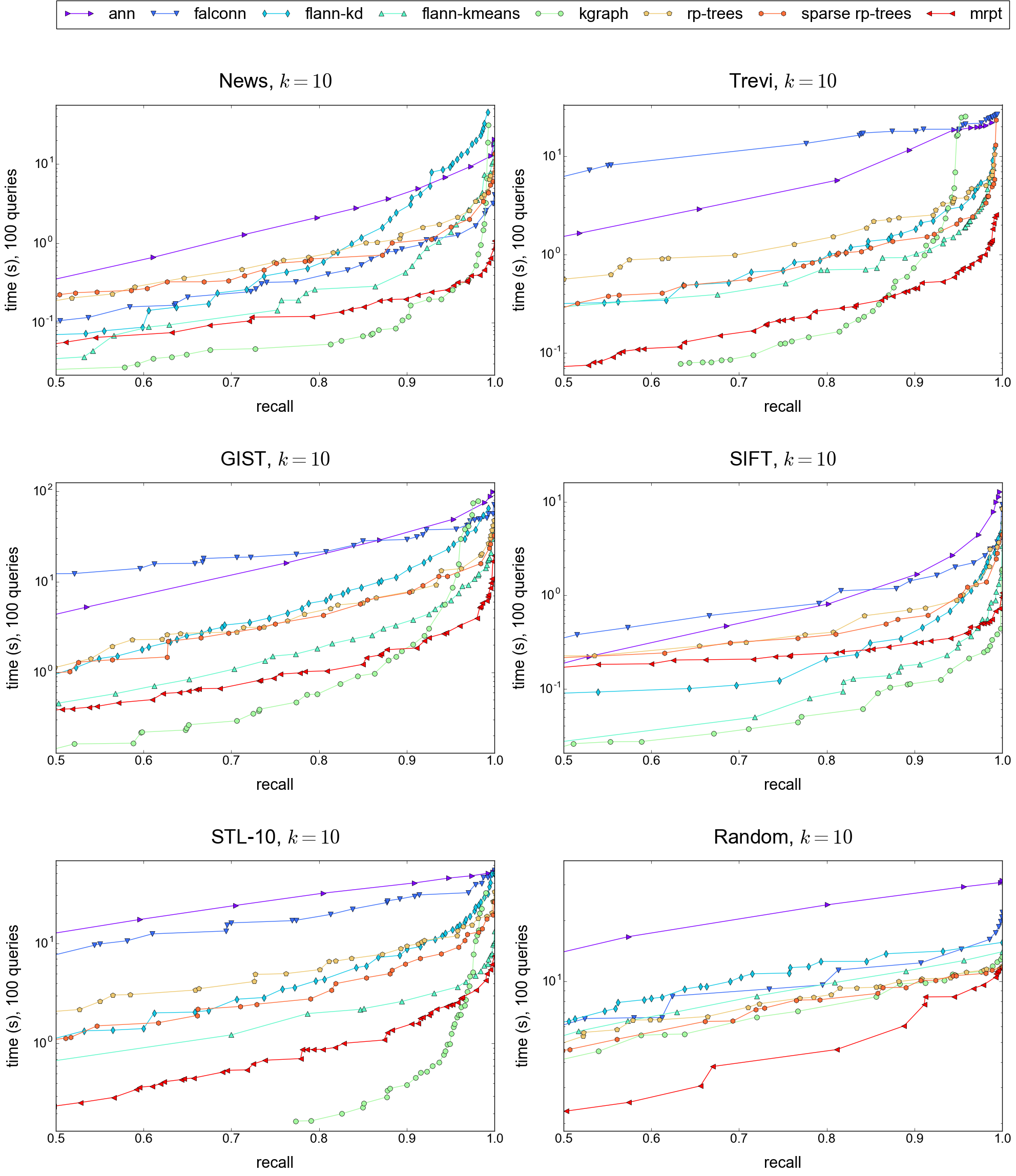}

\caption{Query time (log scale) for 100 queries versus recall on six different data sets with $k = 10$. The compared methods are: k-d tree (ann), multi-probe LSH (falconn), randomized k-d tree (flann-kd), hierarchical k-means tree (flann-kmeans), $k$-NN graph (kgraph) and our method (mrpt). We also include results for our implementation of classic RP-trees (rp trees) and RP-trees with sparsity (sparse rp trees).}
\label{fig:comparison}
\end{figure*}

\begin{table*}[t]
   \centering
   \caption{Query times required to reach at least a given recall level for recall R = 80\%, 90\%, 95\%, 99\%, and for $k=10$. Fastest times for each recall level are emphasized with bold font. MRPT consistently outperforms other algorithms at high recall levels. For example, for R $\ge$ 95\%, 71.4\% of times MPRT achieves the fastest query time (i.e., 10 out of 14) whereas it is only 28.6\% for KGraph (i.e., 4 out of 14).
   ${}^{*)}$ R = 100\% by definition for brute force search.}
   \label{table:results}
\begin{tabular}%{|c|c|c|c|c|c|c|c|c|c|c|}
{ccccccccccc}
    \toprule
    & & 
    \multicolumn{9}{c}{time (s), 100 queries} \\
    \midrule
    data set & R (\%) & ANN & FALCONN & FLANN-kd & FLANN-kmeans & KGraph & RP-trees & MRPT($v=1$) & MRPT & brute force${}^{*)}$ \\
    \midrule
    \multirow{4}{*}{MNIST}
    & 80 & 0.89 & 0.21 & {\bf 0.02} & {\bf 0.02} & {\bf 0.02} & 0.09 & 0.05 & {\bf 0.02} & \multirow{4}{*}{2.59} \\
    & 90 & 1.57 & 0.36 & 0.04 & 0.04 & 0.04 & 0.16 & 0.08 & {\bf 0.03} & \multicolumn{1}{c}{} \\
    & 95 & 2.29 & 0.55 & 0.06 & 0.05 & 0.06 & 0.2 & 0.14 & {\bf 0.04} & \multicolumn{1}{c}{} \\
    & 99 & 3.46 & 1.23 & 0.15 & 0.1 & 0.44 & 0.39 & 0.22 & {\bf 0.07} & \multicolumn{1}{c}{} \\
    \midrule
    \multirow{4}{*}{News}
    & 80 & 2.15 & 0.39 & 0.56 & 0.26 & {\bf 0.05} & 0.71 & 0.64 & 0.12 & \multirow{4}{*}{23.09} \\
    & 90 & 4.42 & 0.88 & 2.92 & 0.45 & {\bf 0.11} & 1.37 & 1.05 & 0.2 & \multicolumn{1}{c}{} \\
    & 95 & 7.35 & 1.23 & 9.55 & 1.39 & {\bf 0.25} & 1.98 & 1.62 & 0.28 & \multicolumn{1}{c}{} \\
    & 99 & 11.83 & 2.58 & 38.81 & 7.5 & 3.24 & 4.3 & 3.87 & {\bf 0.5} & \multicolumn{1}{c}{} \\
    \midrule
    \multirow{4}{*}{GIST}
    & 80 & 20.54 & 21.21 & 6.03 & 1.86 & {\bf 0.59} & 4.77 & 4.19 & 1.03 & \multirow{4}{*}{52.98} \\
    & 90 & 36.18 & 29.19 & 13.35 & 3.63 & 1.87 & 7.69 & 7.54 & {\bf 1.83} & \multicolumn{1}{c}{} \\
    & 95 & 48.13 & 37.92 & 24.41 & 6.22 & 7.94 & 13.6 & 11.89 & {\bf 2.91} & \multicolumn{1}{c}{} \\
    & 99 & 76.57 & 50.29 & 59.12 & 14.14 & - & 24.11 & 20.33 & {\bf 6.1} & \multicolumn{1}{c}{} \\
    \midrule
    \multirow{4}{*}{SIFT}
    & 80 & 0.81 & 0.93 & 0.21 & 0.09 & {\bf 0.05} & 0.4 & 0.38 & 0.24 & \multirow{4}{*}{21.32} \\
    & 90 & 1.67 & 1.47 & 0.41 & 0.18 & {\bf 0.11} & 0.71 & 0.59 & 0.31 & \multicolumn{1}{c}{} \\
    & 95 & 3.1 & 2.06 & 0.84 & 0.31 & {\bf 0.18} & 0.92 & 0.97 & 0.37 & \multicolumn{1}{c}{} \\
    & 99 & 7.88 & 3.2 & 2.81 & 0.86 & {\bf 0.35} & 3.22 & 2.16 & 0.62 & \multicolumn{1}{c}{} \\
    \midrule
    \multirow{4}{*}{Trevi}
    & 80 & 5.5 & 14.62 & 0.97 & 0.7 & {\bf 0.16} & 1.48 & 0.94 & 0.27 & \multirow{4}{*}{22.15} \\
    & 90 & 12.43 & 17.88 & 1.81 & 1.01 & 0.75 & 2.44 & 1.46 & {\bf 0.45} & \multicolumn{1}{c}{} \\
    & 95 & 18.64 & 18.8 & 3.02 & 1.69 & 18.68 & 3.74 & 2.15 & {\bf 0.67} & \multicolumn{1}{c}{} \\
    & 99 & - & 25.54 & 10.54 & 7.61 & - & 8.13 & 5.25 & {\bf 2.07} & \multicolumn{1}{c}{} \\
    \midrule
    \multirow{4}{*}{STL-10}
    & 80 & 31.31 & 18.62 & 4.29 & 2.04 & {\bf 0.18} & 5.46 & 3.06 & 0.88 & \multirow{4}{*}{49.43} \\
    & 90 & 39.38 & 28.62 & 8.76 & 2.72 & {\bf 0.39} & 9.45 & 6.32 & 1.52 & \multicolumn{1}{c}{} \\
    & 95 & 45.0 & 31.46 & 13.39 & 3.53 & {\bf 1.13} & 11.62 & 8.58 & 2.43 & \multicolumn{1}{c}{} \\
    & 99 & 49.67 & 45.42 & 29.96 & 6.1 & 32.12 & 18.56 & 16.89 & {\bf 4.28} & \multicolumn{1}{c}{} \\
    \midrule
    \multirow{4}{*}{Random}
    & 80 & 23.86 & 10.1 & 12.55 & 9.73 & 7.59 & 8.55 & 8.2 & {\bf 4.55} & \multirow{4}{*}{10.9} \\
    & 90 & 27.33 & 12.27 & 13.91 & 11.48 & 9.8 & 10.2 & 9.8 & {\bf 6.9} & \multicolumn{1}{c}{} \\
    & 95 & 29.07 & 14.28 & 14.5 & 12.55 & 10.93 & 10.88 & 10.71 & {\bf 8.6} & \multicolumn{1}{c}{}\\
    & 99 & 30.47 & 17.03 & 15.37 & 13.7 & 12.02 & 11.59 & 11.42 & {\bf 10.36} & \multicolumn{1}{c}{} \\
    \bottomrule
\end{tabular}
\end{table*}



%\bibliographystyle{IEEEtran}

\balance

\bibliographystyle{IEEEtran}
\bibliography{RP_trees_BigData_refs}  

\begin{thebibliography}{10}
\providecommand{\url}[1]{#1}
\csname url@samestyle\endcsname
\providecommand{\newblock}{\relax}
\providecommand{\bibinfo}[2]{#2}
\providecommand{\BIBentrySTDinterwordspacing}{\spaceskip=0pt\relax}
\providecommand{\BIBentryALTinterwordstretchfactor}{4}
\providecommand{\BIBentryALTinterwordspacing}{\spaceskip=\fontdimen2\font plus
\BIBentryALTinterwordstretchfactor\fontdimen3\font minus
  \fontdimen4\font\relax}
\providecommand{\BIBforeignlanguage}[2]{{%
\expandafter\ifx\csname l@#1\endcsname\relax
\typeout{** WARNING: IEEEtran.bst: No hyphenation pattern has been}%
\typeout{** loaded for the language `#1'. Using the pattern for}%
\typeout{** the default language instead.}%
\else
\language=\csname l@#1\endcsname
\fi
#2}}
\providecommand{\BIBdecl}{\relax}
\BIBdecl

\bibitem{Muja2014}
M.~Muja and D.~G. Lowe, ``Scalable nearest neighbor algorithms for high
  dimensional data,'' \emph{Pattern Analysis and Machine Intelligence, IEEE
  Transactions on}, vol.~36, no.~11, pp. 2227--2240, 2014.

\bibitem{Nister2006}
D.~Nister and H.~Stewenius, ``Scalable recognition with a vocabulary tree,'' in
  \emph{2006 IEEE Computer Society Conference on Computer Vision and Pattern
  Recognition (CVPR'06)}, vol.~2.\hskip 1em plus 0.5em minus 0.4em\relax IEEE,
  2006, pp. 2161--2168.

\bibitem{Lowe2004}
D.~G. Lowe, ``Distinctive image features from scale-invariant keypoints,''
  \emph{International journal of computer vision}, vol.~60, no.~2, pp. 91--110,
  2004.

\bibitem{Amit1997}
Y.~Amit and D.~Geman, ``Shape quantization and recognition with randomized
  trees,'' \emph{Neural computation}, vol.~9, no.~7, pp. 1545--1588, 1997.

\bibitem{Hays2007}
J.~Hays and A.~A. Efros, ``Scene completion using millions of photographs,'' in
  \emph{ACM Transactions on Graphics (TOG)}, vol.~26, no.~3.\hskip 1em plus
  0.5em minus 0.4em\relax ACM, 2007, p.~4.

\bibitem{Wang2016}
L.~Wang, S.~Tasoulis, T.~Roos, and J.~Kangasharju, ``Kvasir: Scalable provision
  of semantically relevant web content on big data framework,'' \emph{IEEE
  Transactions on Big Data}, vol. Advance online publication.
  DOI:10.1109/TBDATA.2016.2557348, 2016.

\bibitem{Indyk1998}
P.~Indyk and R.~Motwani, ``Approximate nearest neighbors: towards removing the
  curse of dimensionality,'' in \emph{Proceedings of the thirtieth annual ACM
  symposium on Theory of computing}.\hskip 1em plus 0.5em minus 0.4em\relax
  ACM, 1998, pp. 604--613.

\bibitem{Gionis1999}
A.~Gionis, P.~Indyk, R.~Motwani \emph{et~al.}, ``Similarity search in high
  dimensions via hashing,'' in \emph{VLDB}, vol.~99, no.~6, 1999, pp. 518--529.

\bibitem{Andoni2006}
A.~Andoni and P.~Indyk, ``Near-optimal hashing algorithms for approximate
  nearest neighbor in high dimensions,'' in \emph{Foundations of Computer
  Science, 2006. FOCS'06. 47th Annual IEEE Symposium on}.\hskip 1em plus 0.5em
  minus 0.4em\relax IEEE, 2006, pp. 459--468.

\bibitem{Lv2007}
Q.~Lv, W.~Josephson, Z.~Wang, M.~Charikar, and K.~Li, ``Multi-probe lsh:
  efficient indexing for high-dimensional similarity search,'' in
  \emph{Proceedings of the 33rd international conference on Very large data
  bases}.\hskip 1em plus 0.5em minus 0.4em\relax VLDB Endowment, 2007, pp.
  950--961.

\bibitem{Dong2008}
W.~Dong, Z.~Wang, W.~Josephson, M.~Charikar, and K.~Li, ``Modeling lsh for
  performance tuning,'' in \emph{Proceedings of the 17th ACM conference on
  Information and knowledge management}.\hskip 1em plus 0.5em minus 0.4em\relax
  ACM, 2008, pp. 669--678.

\bibitem{Andoni2015}
A.~Andoni, P.~Indyk, T.~Laarhoven, I.~Razenshteyn, and L.~Schmidt, ``Practical
  and optimal lsh for angular distance,'' in \emph{Advances in Neural
  Information Processing Systems 28}.\hskip 1em plus 0.5em minus 0.4em\relax
  Curran Associates, Inc., 2015, pp. 1225--1233.

\bibitem{Hajebi2011}
K.~Hajebi, Y.~Abbasi-Yadkori, H.~Shahbazi, and H.~Zhang, ``Fast approximate
  nearest-neighbor search with k-nearest neighbor graph,'' in \emph{IJCAI
  Proceedings-International Joint Conference on Artificial Intelligence},
  vol.~22, no.~1, 2011, p. 1312.

\bibitem{Dong2011}
W.~Dong, C.~Moses, and K.~Li, ``Efficient k-nearest neighbor graph construction
  for generic similarity measures,'' in \emph{Proceedings of the 20th
  international conference on World wide web}.\hskip 1em plus 0.5em minus
  0.4em\relax ACM, 2011, pp. 577--586.

\bibitem{Wang2012}
J.~Wang, J.~Wang, G.~Zeng, Z.~Tu, R.~Gan, and S.~Li, ``Scalable k-nn graph
  construction for visual descriptors,'' in \emph{Computer Vision and Pattern
  Recognition (CVPR), 2012 IEEE Conference on}.\hskip 1em plus 0.5em minus
  0.4em\relax IEEE, 2012, pp. 1106--1113.

\bibitem{Bentley1975}
J.~L. Bentley, ``Multidimensional binary search trees used for associative
  searching,'' \emph{Communications of the ACM}, vol.~18, no.~9, pp. 509--517,
  1975.

\bibitem{Arya1998}
S.~Arya, D.~M. Mount, N.~S. Netanyahu, R.~Silverman, and A.~Y. Wu, ``An optimal
  algorithm for approximate nearest neighbor searching fixed dimensions,''
  \emph{Journal of the ACM (JACM)}, vol.~45, no.~6, pp. 891--923, 1998.

\bibitem{Fukunaga1975}
K.~Fukunaga and P.~M. Narendra, ``A branch and bound algorithm for computing
  k-nearest neighbors,'' \emph{IEEE transactions on computers}, vol. 100,
  no.~7, pp. 750--753, 1975.

\bibitem{Beygelzimer2006}
A.~Beygelzimer, S.~Kakade, and J.~Langford, ``Cover trees for nearest
  neighbor,'' in \emph{Proceedings of the 23rd international conference on
  Machine learning}.\hskip 1em plus 0.5em minus 0.4em\relax ACM, 2006, pp.
  97--104.

\bibitem{Yianilos1993}
P.~N. Yianilos, ``Data structures and algorithms for nearest neighbor search in
  general metric spaces,'' in \emph{SODA}, vol.~93, no. 194, 1993, pp. 311--21.

\bibitem{Leibe2006}
B.~Leibe, K.~Mikolajczyk, and B.~Schiele, ``Efficient clustering and matching
  for object class recognition.'' in \emph{BMVC}, 2006, pp. 789--798.

\bibitem{Silpa2008}
C.~Silpa-Anan and R.~Hartley, ``Optimised kd-trees for fast image descriptor
  matching,'' in \emph{Computer Vision and Pattern Recognition, 2008. CVPR
  2008. IEEE Conference on}.\hskip 1em plus 0.5em minus 0.4em\relax IEEE, 2008,
  pp. 1--8.

\bibitem{Muja2009}
M.~Muja and D.~G. Lowe, ``Fast approximate nearest neighbors with automatic
  algorithm configuration.'' \emph{VISAPP (1)}, vol.~2, pp. 331--340, 2009.

\bibitem{Dasgupta2009}
S.~Dasgupta and Y.~Freund, ``Random projection trees for vector quantization,''
  \emph{IEEE Transactions on Information Theory}, vol.~55, no.~7, pp.
  3229--3242, 2009.

\bibitem{Dasgupta2013}
S.~Dasgupta and K.~Sinha, ``Randomized partition trees for nearest neighbor
  search,'' \emph{Algorithmica}, vol.~72, no.~1, pp. 237--263, 2015.

\bibitem{Achlioptas2001}
D.~Achlioptas, ``Database-friendly random projections,'' in \emph{Proceedings
  of the twentieth ACM SIGMOD-SIGACT-SIGART symposium on Principles of database
  systems}.\hskip 1em plus 0.5em minus 0.4em\relax ACM, 2001, pp. 274--281.

\bibitem{Li2006}
P.~Li, T.~J. Hastie, and K.~W. Church, ``Very sparse random projections,'' in
  \emph{Proceedings of the 12th ACM SIGKDD international conference on
  Knowledge discovery and data mining}.\hskip 1em plus 0.5em minus 0.4em\relax
  ACM, 2006, pp. 287--296.

\bibitem{Jegou2011A}
H.~J{\'e}gou, M.~Douze, and C.~Schmid, ``Product quantization for nearest
  neighbor search,'' \emph{IEEE Transactions on Pattern Analysis and Machine
  Intelligence}, vol.~33, no.~1, pp. 117--128, 2011.

\bibitem{Jegou2011B}
H.~Jegou, R.~Tavenard, M.~Douze, and L.~Amsaleg, ``Searching in one billion
  vectors: Re-rank with source coding,'' in \emph{Proceedings of the {IEEE}
  International Conference on Acoustics, Speech, and Signal Processing,
  {ICASSP} 2011}, 2011, pp. 861--864.

\bibitem{Winder2007}
S.~A.~J. Winder and M.~A. Brown, ``Learning local image descriptors,'' in
  \emph{2007 {IEEE} Computer Society Conference on Computer Vision and Pattern
  Recognition ({CVPR} 2007)}, 2007.

\bibitem{Coates2011}
A.~Coates, A.~Y. Ng, and H.~Lee, ``An analysis of single-layer networks in
  unsupervised feature learning,'' in \emph{Proceedings of the Fourteenth
  International Conference on Artificial Intelligence and Statistics, {AISTATS}
  2011}, 2011, pp. 215--223.

\end{thebibliography}

\end{document}